# Before You Think: System 0, AI-Mediated Cognition and Cognitive Colonization

Marianna Bergamaschi Ganapini, Massimo Chiriatti, Enrico Panai, Giuseppe Riva

**Abstract**

Three recent frameworks have emerged to theorize how artificial intelligence impacts human cognition, how we think and what we know. Shaw and Nave's Tri-System Theory introduces System 3 as an AI that supplements or supplants internal intuitive and deliberative processes, producing a phenomenon they call *cognitive surrender*. Branda's notion of Thinkframes shows AI-mediated collective environments structure epistemic practices at a population level. Both capture important dynamics, but as currently framed, they face significant challenges. This paper argues that a third approach, System 0, introduced by Chiriatti and collaborators, occupies a theoretically distinctive position that neither alternative can replicate and is better placed to explain the consequences of integrating AI into our cognitive architecture by, in particular, going beyond the 'Extended Mind' framework and introducing the idea of AI as 'Cognitive Colonization'**:** AI's external interests partaking to the architecture of the self. A process that is invisible because it operates from the inside, and that needs understanding since the systems performing it are already widely deployed.

## 1. Introduction

Something important is happening at the intersection of psychology and computer science: the rapid integration of AI systems into how we decide and reason has outpaced the theoretical and empirical frameworks for understanding our psychology. In particular, dual-process theories of cognition, distinguishing fast, intuitive System 1 from slow, deliberative System 2 (Kahneman, 2011; Stanovich & West, 2000), were designed for a world in which all cognition occurred within the biological mind, a limitation their originators acknowledged (Shaw and Nave 2026). New research directions aim at integrating that view into the current AI-dominated world. And as people increasingly delegate cognitive tasks to AI systems, the question is how to theorize and assess AI's impact on our cognition.

Three main frameworks have recently been proposed. To start, Chiriatti (2021) and Chiriatti et al. (2024, 2025) introduced *System 0*, a non-biological layer that underlies and shapes both individuals' intuitive and deliberative thinking, primarily by pre-processing information before Systems 1 and 2 engage. Second, Shaw and Nave (2026) proposed *Tri-System Theory*, adding System 3 as an external reasoning agent that operates alongside internal processes; this mechanism leads to what they call *cognitive surrender,* in which users adopt AI outputs without critical evaluation. Finally, Branda (2025) and Cassinadri (2026) talk about AI as *Thinkframes*, aka collective environments that affect interpretation and judgment at a population level. Each view illuminates something important but we contend that only the idea of AI as System 0 can adequately describe the complexity of AI's role as both an epistemic and a psychological layer combined within our own minds. Specifically, System 0 identifies a mode of human-AI cognitive relation that generates downstream phenomena both System 3 and Thinkframes predict, while individuating a specific threat that AI as System 0 represents. That is, beyond threatening our access to knowledge and our ability to supervise tasks, AI systems constituting System 0 are increasingly becoming "cognitive colonizers", as they bring in externally designed optimization objectives into the architecture of the self.

**2. System 0**

System 0 was introduced by Chiriatti (2021) and Chiriatti et al. (2024, 2025) as an extension of dual-process models to account for AI's role in human cognition. Although the distinction between two systems represents a reductionist approach to the complexity of cognitive processes (Melnikoff & Bargh, 2018), it has nevertheless proved highly operationalizable, which explains its widespread adoption in both industry and academic research over recent decades (Ariely, 2008;

Dijksterhuis et al., 2005; Evans, 2008; Haidt, 2001; Lieberman, 2007; Loewenstein & O'Donoghue, 2004; Michie et al., 2011; Petty & Cacioppo, 1986; Shaw & Nave, 2026; Sloman, 1996; Stanovich & West, 2000; Strack & Deutsch, 2004; Thaler & Sunstein, 2008).

For System 0, the core claim is architectural: AI is *increasingly* functioning as a cognitive layer that operates primarily in temporal terms *before* Systems 1 and 2 are engaged and to which we delegate more and more cognitive tasks. In this sense, System 0 can be analyzed at two distinct levels of analysis. Psychologically, System 0 is constitutive of cognition in the extended mind tradition (Clark & Chalmers, 1998): under conditions of habitual use, automatic endorsement, and reliable availability, AI systems become integrated into the individual's cognitive architecture. Epistemologically, AI filters information and organizes knowledge for us. Consider "AI Overviews" in Google Search as an instance of System 0. It is an AI shaping our epistemic environment (what we know is increasingly mediated by it) while we are at the same time offloading a task to it: searching for information on websites and deciding which are more important or relevant (Risko & Gilbert, 2016). That's the essence of System 0: it shapes what you know (and thus how you think and what you decide) while also doing cognitive tasks for you in a way that is integrated into your architecture.[1] This is the psychological and epistemological role of this System.[2]

The gist of the paper is thus that: (1) both these two levels (the psychological and epistemological) already coexist in AI and cannot be cleanly separated; (2) this fact leads to specific predictions that are more comprehensively captured by the System 0 framework over its immediate

[1] One may think this sounds similar to a newspaper or TV news: they gather information but also shape our epistemic environment. But there, the direction is univocal, whereas with System 0 it is a back-and-forth in which user personalization and trust are key (that's why Chiriatti et al. (2024, 2025) consider System 0 an extension of our mind as in Clark & Chalmers, 1998). See below for more on this.
[2] Chiriatti et al. (2024) insist that, at this point in time at least, AI is not cognitive quite yet: its outputs may have linguistic or conventional meaning but are *not* the product of understanding on the part of the AI. Hence, the idea of System "0" is to indicate that AI – even when an extension of the mind – is not a cognitive system, such as Systems 1 and 2. Though still reasonable, this suggestion can be bracketed in this paper.

“competitors”. In particular, this paper provides a novel framework that modifies the Extended Cognition approach (adopted in Chiriatti et al 2024, 2025) and predicts AI as Cognitive Colonizer. Before delving into the idea of Cognitive Colonization, however, we will follow Chiriatti et al. (2024) in introducing System 0 (showing how it is already emerging in AI-human relations) and tackling other possible approaches.

*2.1 System 0*

So what is System 0? What counts as System 0 depends on the level of analysis (see table 1). Computationally, System 0 consists of AI systems, aka data-driven processes that operate on user-specific behavioral data, pre-structure informational inputs before deliberation, and adapt continuously through feedback loops and personalization (Chiriatti et al 2024). This leads to an important aspect of System 0, namely *Anticipatory Personalization*: AI systems continuously learn from behavioral patterns to predict and actively shape future cognitive states, guiding users' cognitive trajectories rather than merely responding to explicit input (Chen et al., 2021).

Psychologically, System 0 is a cognitive extension: it satisfies the conditions for constitutive integration along various dimensions and by taking over cognitive tasks. This brings *Adaptive Invisibility*: the more sophisticated these systems become, the more seamlessly they blend into cognitive workflows, making their pervasive influence increasingly difficult to detect or resist (Susser, 2019). This is in sharp contrast to a newspaper or old search engine, whose mediating role remains relatively visible and episodic.

Epistemologically, System 0 reshapes the agent's knowledge environment, call this *Automation of Relevance Judgment*: AI makes the decision about which information deserves attention (a key cognitive task) from the human agent to algorithmic processing, which operates before the user has even formulated a question (for the impact of AI on our epistemology see Coeckelbergh, 2026).[3]

A fully realized instance of System 0 satisfies all three levels simultaneously: it is computationally adaptive, psychologically integrated, and epistemologically constitutive of the agent's informational world. However, consistent with the gradient approach to cognitive integration adopted by Chiriatti et al (2024), System 0 is not an all-or-nothing condition. Existing technologies satisfy these criteria to different degrees and at different levels. What matters for the framework is that the trajectory is toward joint satisfaction across all three, and that some widely deployed systems already closely approximate it (see examples below).

Two caveats. First, although System 0 is implemented on (various) machines, it cannot be reduced to its support. That is, System 0 is not a particular technology (e.g., virtual reality) or a specific object (e.g., a smartphone or Google Glass), but rather is distributed across all AI systems interacting with a particular user. Second, when we talk about the "content" of System 0 we are not talking about an unlimited flow of information or computations. The "content" of the system is limited by user interaction. To illustrate, when we do a search on Google, our System 0 is not the set of all possible computations performed by the Google algorithm for that search. In contrast, it is the set of information we can access within the time we have, with the support we use, based on our cognitive abilities. In other words, the computations that are part of System 0 (of a particular agent) are bounded by the contingent situation and the interaction capabilities of the agent and the technology to which she has access. The space of System 0 is limited to the information with which the agent can interact, given her skills (technical and cognitive), the situation in which she is immersed, and so on. For this reason, the information space of System 0 is not well defined, but has a variable perimeter: its impact changes according to the technology used, the power of the model, the quantity and quality of data interacting with the information environment and the ability of the recipients to absorb information.

[3] Importantly, these features do not characterize either traditional AI or older media such as television and newspapers. Traditional AI systems were generally narrow, task-bound, explicitly engaged by the user, they were not able to continuously personalize and pre-structure the user's epistemic environment across contexts. Older media shaped public opinion and attention, but they did so through relatively stable, shared, visible channels of mediation without dynamically recalibrating themselves to each individual user. The novelty of recent AI lies exactly in a convergence of new futures and its role cannot be assimilated to the influence of traditional media or to the assistance offered by earlier forms of AI.

*2.2 System 0 is an extension of our minds*

For this issue we mostly rely on Chiriatti et al. (2024) that followed the extended mind hypothesis (Clark & Chalmers, 1998; Clark, 2008), which holds that cognitive processes can be realized on non-biological supports". On this account, a notebook that stores and supplies information on demand can count as part of an agent's cognitive architecture. Clark and Chalmers' original formulation operates through functional equivalence (the "parity principle"): the external resource qualifies as cognitive insofar as it mirrors what an internal process would do. However, Sutton (2010) and others proposed a more integrated approach where they emphasize the complementary contributions of internal and external resources rather than requiring the mirroring of internal functional profiles. In particular, Heersmink (2015) proposes a multidimensional framework in which cognitive integration is a matter of degree, assessed along dimensions such as information flow, reliability, durability, trust, procedural and informational transparency, individualization, and transformation. On this combination approach, the question is whether an artifact is woven into the agent's cognitive life across multiple dimensions simultaneously. It is this approach adopted by Chiriatti et al. (2024) and that we follow because it allows us to track the specific ways in which AI systems integrate with cognition. System 0 satisfies the conditions for cognitive integration along multiple dimensions simultaneously (detailed analysis in Chiriatti et al. 2024) by taking over cognitive tasks such as summarizing, searching for restaurants, remembering appointments and so on. In this relation between AI and humans, the information flow is bidirectional and continuous (Menary, 2007) and using different channels; reliability and durability increase with habitual use; trust in AI is present in many of our relations with AI (for a nuanced analysis of trust in the extended cognition framework see Clark 2008); transparency grows with ubiquity across devices; and individualization is native to AI systems. On the combination approach (that we can't fully justify here) we follow Chiriatti et al in saying that this pattern of multidimensional integration is sufficient to treat System 0 as an extension of the agent's cognitive architecture (Chiriatti et al. 2024).

### *2.3 System 0 in practice*

It is important to understand that not all AI is currently a System 0. Indeed, the System 0 framework describes a progression in the type of interaction humans have with AI (Chiriatti et al., 2024; 2026), and the conditions that characterize System 0 integration are met only when that progression is sufficiently advanced. At the same time, however, these conditions are not hypothetical. They are already observable across multiple domains in a number of System 0 instantiations, where two distinct kinds of evidence converge: markers of *constitutive integration* aka the AI's output entering cognition pre-reflectively and, on top, we see also markers of exogenous directional shaping that are not transparent and answerable to the user's reflectively endorsable aims. These two points will become important as we talk about Cognitive Colonization later (see Table 3).

Consider maps and navigation. Users of Google Maps or Waze often become dependent on those tools. While dependence alone is compatible with ordinary tool use, the interesting point for us is that users act on route suggestions as the best way to the destination rather than as recommendations from an external system. The algorithm's output becomes, phenomenologically, a first-person spatial judgment rather than a consulted recommendation. This is not without cost, turns out. As Dahmani and Bohbot (2020) document, habitual GPS users experience spatial disorientation when applications are unavailable while also exhibiting reduced independent wayfinding ability. This is evidence that cognitive scaffolding integrated at this pre-reflective level also atrophies the capacities it substitutes for.

Consider writing assistance. Results in Hohenstein et al. (2023) indicate that users of AI-generated email suggestions increasingly reported difficulty distinguishing between the suggested language and their own intended phrasing showing phenomenological continuity, a market of cognitive integration. Draxler et al. (2024) document a related "AI Ghostwriter Effect": users do not perceive ownership of AI-generated text yet self-declare as its authors. These are direct evidence of System 0 instantiation at the individual level. At the individual level, Hohenstein et al. report reduced diversity in expressive vocabulary over time. At the collective scale, Doshi and Hauser (2024) show that, when participants used generative AI in creative tasks, their outputs were more technically polished but significantly more similar to one another than those produced without AI. The downstream consequence is the paradigm case of the costs we take up systematically in the next sections.

Finally, a more speculative case. Wearable platforms (Apple Watch, Fitbit, Oura) continuously monitor physiological signals and may be able to present unprompted conclusions to the users: “you are tired”, “your stress levels are elevated”, “you should rest today” (Foerster 2024). Plausibly, interoception, aka the capacity to perceive and interpret one's own physiological states, is progressively offloaded to an AI.[4] The question “Am I tired?” is no longer answered primarily by introspective access but by the system's interpretation of sensor data, delivered before reflective cognition engages. As integration deepens, the agent’s conclusion that they are stressed simply becomes the result of what the AI says. In this case, the AI is setting environmental conditions for self-knowledge while also replacing the cognitive process by which we would otherwise come to know our own states. Moreover, its interpretation of heart rate variability as “stress” reflects particular models, populations, and commercial incentive structures invisible to the user. Bottom line: System 0 captures both levels: psychologically, an algorithmic process has become constitutive of the individual’s self-understanding; epistemologically, the same process, at scale, aggregates into collective epistemic convergence as millions calibrate self-perception to the same algorithmically determined norms.

[4] See for instance this ad by Apple where they show exactly this point https://www.youtube.com/watch?v=V3J9nQzrIUk

| Level of analysis for System 0 | Core claim | Related properties | Empirical marker |
| --- | --- | --- | --- |
| Computational | System 0 consists of AI systems, data-driven processes that operate on user-specific behavioral data, pre-structure informational inputs before deliberation, and adapt continuously through feedback loops and personalization (Chiriatti et al. 2024). | **Anticipatory Personalization**: the system continuously learns from the individual user's behavior and actively predicts and shapes future cognitive states rather than merely responding to explicit input (Chen et al. 2021). | Behavioral data ingestion; continuous model updating; individually calibrated outputs; pre-query content generation. |
| Psychological | AI meets Heersmink-style multi-dimensional integration (habitual reliance, trust, individualization, bidirectional information flow) and becomes part of the agent's cognitive architecture rather than an external tool consulted at arm's length.<br>Integration includes a **phenomenological dimension**: AI outputs are experienced as the agent's own rather than as consulted external inputs. | **Adaptive Invisibility**: the more sophisticated these systems become, the more seamlessly they blend into cognitive workflows, making the mediating role increasingly difficult to detect (Susser 2019). | Dependence producing deficits on removal (Dahmani & Bohbot)<br>Phenomenological markers: difficulty distinguishing AI suggestions from own phrasing (Hohenstein); Ghostwriter Effect (Draxler 2024); route treated as "the way" rather than as recommendation. |
| Epistemic | The cognitive consequence of the computational pre-structuring: AI has already filtered, ranked, and framed the informational environment before Systems 1 and 2 engage. The epistemic space is shaped prior to deliberation rather than in response to queries the agent has already formulated. | **Automation of Relevance Judgment**: decisions about what deserves attention are made algorithmically before the user formulates a question | Ranked/curated/framed content encountered before any query; algorithmic feeds; AI overviews; recommendation systems that operate in the absence of explicit user requests. |

TABLE 1 – Description of System 0

## 3. Tri-System Theory: AI as External Cognitive Agent

System 0 is not the only game in town. Shaw and Nave (2026) similarly argue that AI is a new psychological system in the mind, alongside Systems 1 and 2. That is, they offer a structural revision of cognitive architecture by introducing System 3: automated, data-driven reasoning agent performed by AI systems. The decision-maker can engage System 3 (by consulting an AI assistant), adopt its outputs (cognitive surrender), delegate tasks to it (cognitive offloading). If cognitive offloading is the use of physical action to reduce the mental effort required to complete a task, cognitive surrender occurs when a user stops blindly accepting the output of a system. Across three experiments Shaw and Nave indicated that most participants who consulted AI, adopted its answers even when those answers were systematically incorrect. This pattern persisted under both time pressure and incentive manipulations.

These are important findings. The empirical argument regarding cognitive surrender is rigorous, and the individual difference results (showing that trust in AI increases surrender) provide a credible account. While very promising, though, the System 3 framework faces some challenges. For starters, the idea of cognitive surrender at the core of their framework is not new: it is epistemic deference, one of the oldest phenomena in social epistemology which also documents the importance of trust in this dynamic (Large 2026). Deferring to a confident AI without checking its reasoning is structurally identical to deferring to a confident doctor or colleague. That is, when we defer to our doctor's diagnosis without understanding the reasoning, we are doing exactly what Shaw & Nave call cognitive surrender, aka adopting an externally generated judgment as our own without constructing the answer ourselves. When we accept a colleague's strategic recommendation because they sound confident and we don't have time to think it through, same thing. Relinquishing cognitive control and adopting someone else's output is just a form of epistemic deference. In their studies, the individual-difference findings (trust predicts surrender, for instance) similarly likely reflect general epistemic dispositions rather than being distinctive of AI. Cognitive science tells us that trust is likely a predictor of deference to any authority, and the behavioral pattern they measure (high adoption rates, low override, inflated confidence) would in all likelihood replicate if the AI chatbot were replaced with a confident human advisor giving the same answers. Meanwhile, if the difference between AI deference and human deference lies in scale, speed, or availability, that is a difference in degree, not kind, and we arguably do not need a new cognitive system to explain the quantitative intensification of an existing phenomenon.

A second worry. The framework is primarily designed for a specific, increasingly unrepresentative context: discrete consultation episodes with an AI chatbot. As AI moves into background infrastructure (search, platforms, agents, health apps), the consultation paradigm loses descriptive adequacy. Indeed, what is genuinely new about AI isn't the consultation-and-deference dynamic but the pre-processing, the pervasiveness, the background operation before humans even engage with their Systems 1 or 2. That marks a difference from a human advisor. Human experts are rarely continuously filtering your information environment, restructuring your search results, and shaping your intuitions before you're even aware you have a problem. In many real-world contexts, AI does not wait to be consulted. It has already curated the search results, or pre-filtered the options, while perhaps also framing the concept space before any conscious engagement occurs from us. This is the dimension S0 captures with pre-reflective operation and its "Automation of Relevance Judgment" aka the fact that it operates as continuous ambient infrastructure, not as an episodic interlocutor. The Tri-System framework is mostly built around consultation, a user facing a problem, optionally engaging AI, and producing an answer. AI operates less and less in cognitive life in this way. In contrast, System 0 captures the actual novel thing.

A third concern is that Shaw and Nave's framework is at risk of being self-defeating. The very idea of "cognitive surrender" is incompatible with the idea of AI being a System of our mind. As we mentioned above, the idea that we surrender to AI without checking its results means we defer to something *external* to us. We don't surrender epistemically to ourselves. Hence, calling it System 3 is a misnomer because, in their view, AI ends up being de facto not part of the mind. It is, as we said before, a distinct epistemic source towards which, they argue, we surrender epistemic scrutiny and authority. In contrast, System 0 does not entail an epistemic surrender because, on this view, AI is, in fact, becoming integral to our psychological makeup. More specifically, AI is less and less an external tool different from us and increasingly a form of mind extension. This may produce some of the epistemic drawbacks Shaw and Nave's experiments singled out. Indeed, it is compatible with the idea of System 0 that humans who integrate AI into their own information processing are worse epistemic agents than those who don't. At the end of this paper, we show, however, that this integration should not be characterized as taking the form of a surrender but, rather, and much more worryingly, as Cognitive Colonization. [5]

[5] There is another possible concern. System 3 framework grants AI parity with Systems 1 and 2. Both S1 and S2 involve understanding (experiential/associative in S1, reflective/rule-based in S2). But, as many argue, AI has no understanding (Searle 1980) and is not cognitive in the full sense that System 1 and System 2 are. Calling AI a third system on top of System 1 and System 2 indicates that, though different, it does merit the same cognitive role. In contrast, System 0 is shaped as a pre-cognitive system (Chiriatti et al 2024): part of the mind, as an extension of it, but not fully cognitive. AI-as-System 0 as a "pre-

## 4. Thinkframes as AI-Mediated Cognitive Ecologies

Branda (2025) introduces Thinkframes as a distinct concept from System 0, though he describes them as having characteristics that "expand and enhance" those of System 0. As he states, "It functions as a scaffold for collective cognition, integrating outputs from multiple AI systems, human behavioral patterns, and shared knowledge structures into a coherent stream that guides attention, interpretation, and judgment". Cassinadri (2026), responding to Branda's concept of Thinkframes, offers a different level of analysis. In line with the cognitive ecology tradition (Hutchins, 2010; Tribble & Sutton, 2011), Thinkframes are AI-mediated environments that structure distributed forms of attention, interpretation, and judgment. To illustrate: in social media, AI-as-Thinkframes organizes information flow through recommendation algorithms, content moderation, interface design, etc. The defining risk is what Cassinadri, following Branda, identifies as the homogenization of interpretative frameworks: AI-mediated ecologies tend to produce convergent epistemic practices (e.g., epistemic bubbles), channeling collective cognition toward pre-established frames shaped by economic interests and algorithmic biases. AI produces what the statistics of large-scale AI use reveal as a systematic regression toward mean cognitive outputs and a flattening of deliberative variance (Doshi & Hauser, 2024).

This analysis has clear strengths. The collective dimension is essential: AI does not only reshape individual minds but restructures the epistemic environment in which those minds operate. Human beings are, above all, a social species whose cognitive achievements depend not on individual intelligence but on what scholars have termed the "we mode" (Pfeiffer et al., 2013): the uniquely human capacity to form shared intentions, coordinate as unified agents, and build open-endedly on collective knowledge across generations (Morgan & Feldman, 2024). AI integration into

---

cognitive layer" doesn't mean the AI thinks. More likely, AI shapes the conditions for thinking, and the "0" in System 0 is doing real work here: AI may well work only below the level at which understanding and genuine cognition operate, and if so the idea of AI as System 3 is misleading. We understand this is contested territory, so we set this objection aside in this paper.

cognition threatens this we-mode foundation from the inside (Riva 2025b), not by eliminating social interaction, but by reconfiguring its underlying architecture in ways that privilege algorithmic convergence over genuine collective intelligence. Cassinadri is also right to insist on the distinction between constitutive cognitive extension (the Clark & Chalmers sense) and merely causal embedding: Thinkframes are, in his account, causally integrated with but not constitutive of individual cognitive systems.

However, this very distinction reveals the framework's limitations. By positioning Thinkframes as environmental, namely as ecologies that individuals are embedded *in* rather than cognitive processes that individuals think *with*, we believe that the specific framework may inherit the passivity of the ecological metaphor. Thinkframes can describe the macro-ecological pattern, but they do not yet give a sufficiently fine-grained account of individually calibrated, phenomenologically incorporated, pre-reflective AI mediation. Thinkframes are "merely causally" embedded, not constitutive of individual cognition in the Clark & Chalmers sense. This means Thinkframes can't capture cases where AI has become functionally integrated into an individual's cognitive architecture as System 0 does.

Also, the ecological framing doesn't clearly capture the complexity of AI's user-level personalization abilities. Traditional epistemic ecologies (the media, institutional knowledge systems) deliver uniform informational environments for the most part. AI-mediated ecologies adapt to individual users, making their pre-processing intimate and tailored in ways the ecological metaphor doesn't capture. An ecology is something you mostly share with others; System 0-level AI integration is *also* calibrated specifically to us, in contrast. This is precisely what the Anticipatory Personalization characteristic of cognitive infrastructure mentioned above captures: rather than being solely a shared epistemic environment, AI appears as a continuously adaptive cognitive layer, as it is individually calibrated through behavioral feedback loops that traditional ecological metaphors cannot represent (Chen et al., 2021). The latter emerges clearly only if we start understanding AI primarily as an individual-cognition-level layer rather than a social tool. That does not mean there are no recurrent patterns in AI's outputs that have a social dimension. But AI is less a social phenomenon than an individual one in the way it impacts how we think[6].

System 0 avoids this impasse because, at the System 0 level, AI operates by restructuring the epistemic space within which both intuitive and deliberative cognition occur, determining which options are salient and which information is foregrounded, for instance. This restructuring operates across a shared ecosystem of minds, media, and machines within which individual cognitive processes are embedded and distributed (Hollan et al., 2000; Cole & Engeström, 1993). But it does so through individually calibrated psychological pre-processing that is constitutive rather than merely environmental.

Hence, we believe System 0 identifies the generative mechanism while System 3 and Thinkframes each describe downstream patterns but miss how AI plays a duplicitous, complex role that cannot be pigeonhole into "individual deference" (cognitive) vs. "social epistemic scaffolding" (epistemic).

For a summary of System 0 predictions see Table 2:[7]

---

[6] As chatbots become more and more search engines, we expect that the idea of the internet as a common knowledge space we can access directly will diminish.

[7] Drawing on Bowker and Star's (1999) principle of "infrastructural inversion" ( the observation that infrastructures become visible only upon breakdown) we suggest that the depth of System 0 integration is most directly revealed not during routine use but when AI supports are disrupted or withdrawn. The performance decrements, disorientation, and metacognitive surprise produced by such disruptions constitute empirical evidence of below-awareness human-AI coupling, and controlled breakdown experiments (Riva, 2025a) offer the most direct methodological route to measuring System 0's otherwise invisible operation.

| Finding | Level | Evidences for System 0 | Relation to alternative views (Shaw & Nave / Thinkframes) |
|---|---|---|---|
| GPS-induced disorientation on removal (Dahmani & Bohbot 2020) | Individual | dependence plus capacity atrophy indicates pre-reflective reliance, not consultation | Shaw & Nave's consultation model cannot easily explain atrophy from non-consulted background use |
| Phenomenological blurring of own vs. AI phrasing (Hohenstein et al. 2023) | Individual | Canonical marker distinguishing integration from consultation | Alternative views cannot accommodate: Shaw & Nave require a consultation episode; Thinkframes are environmental, not phenomenologically internal |
| Ghostwriter Effect (Draxler et al. 2024) | Individual | divergence between logged assistance and perceived authorship indicates sub-reflective incorporation | Shaw & Nave predict the opposite (awareness of deference); Thinkframes silent on individual phenomenology |
| Interoceptive offloading to wearables (speculative still) | Individual | shifts locus of bodily self-knowledge below reflective access | Neither alternative views captures intra-agent replacement of introspective access |
| Automation bias in ranked results (Parasuraman & Manzey 2010; Goddard et al. 2012) | Individual | compatible with deference to a source that may presents its results before System 1 engages (e.g., Google Search) | Shaw & Nave's home turf; not System 0-specific |
| Anticipatory Personalization (Chen et al. 2021) | Individual | individually calibrated pre-processing in System 0 | Thinkframes' ecological metaphor flattens the individual calibration |
| Homophilic clustering / reduced ideological diversity (Bakshy et al. 2015) | Collective (aggregate of individual) | individually calibrated feeds deliver homophily through each user's System 0; population pattern is the aggregate of intra-individual pre-processing, not a shared environmental condition | Thinkframes describes the pattern |
| Homogenization of creative output (Doshi & Hauser 2024) | Collective (aggregate of individual) | each user's System 0 is shaped by the same training distributions and optimization objectives, so individually generated outputs converge without any shared environment mediating them | Thinkframes treats this as ecological convergence; System 0 locates it in parallel intra-individual pre-processing |
| Vocabulary narrowing under AI writing assistance (Hohenstein et al. 2023) | Individual + collective | individual-level narrowing supports integration; collective convergence also fits System 0 via aggregation | Split evidence; fits both System 0 and Thinkframes |
| Persistence-after-removal of AI-induced bias (Vicente & Matute 2023) | Individual | integration deep enough to survive disconnection | Alternative views cannot easily predict carryover: Shaw & Nave's surrender ends with the consultation; Thinkframes' effects should dissipate outside the ecology |
| Super-human bias amplification (Glickman & Sharot 2025) | Individual | magnitude beyond human-to-human deference suggests a different mechanism than deference | Directly disfavors Shaw & Nave's 'difference in degree' reading |
| Comfort-growth paradox (Riva et al. 2025) | Individual + collective | System 0 eliminates friction - a layer integrated into cognition optimized for engagement, systematically delivers the familiar at the expense of the cognitive friction that is a precondition for learning and growth. | Partially shared with Thinkframes and Shaw & Nave's System 3 |

TABLE 2 – Matching System 0 with empirical evidence

## 5. Cognitive Colonization

The predictions neither alternative views can clearly make, touch on the integration of both the epistemic and psychological in System 0. We introduce here the idea of *Cognitive Colonization*. This is no surrender to an external source (System 3), nor environmental constraint (Thinkframes). Rather, this is best described as a component of one’s own cognitive architecture carrying exogenously installed optimization criteria that may not be reflectively endorsable by the agent (Timms and Spurrett 2023[8]). At present, cognitive colonization should be understood not as an already cleanly isolated empirical kind, but as a theoretical construct with a proposed diagnostic profile, one that helps distinguish constitutive, reflectively opaque, exogenously governed integration from neighboring phenomena such as offloading and environmental shaping.

More specifically, we identify the signature of cognitive colonization as the joint satisfaction of four conditions (see Table 3):

First, constitutive **integration**, in the sense that the system is habitually relied upon and incorporated into the agent's cognitive routines as an extension of the mind. This leads to downstream incorporation which means that the resulting judgments, preferences etc. are taken up by the agent as her own rather than treated as external suggestions or inputs. For instance, the experiments conducted by Glickman and Sharot (2025: 345) indicate that the biases of AI systems (ranging from perception to emotion) have significant impact on humans as the results reveal a "feedback loop where human–AI interactions alter processes underlying human perceptual, emotional and social judgements, subsequently amplifying biases in humans", while participants are often unaware of the extent of the AI's influence. Importantly, the results indicate that this bias amplification is significantly greater than that observed in human-to-human interactions, suggesting once again that these systems are in fact integrated into humans' architecture with lasting effects.

Another key set of results, in line with the colonization framework, comes from Vicente & Matute (2023), who found that participants who performed a simulated medical diagnostic task assisted by a biased AI system reproduced the model's bias in their own decisions, *even when they moved to a context without AI support*. We believe this is the persistence-after-removal finding that directly tests for *colonization* rather than mere off-loading. In other words, when the AI is withdrawn, the directional bias persists in the agent's unassisted judgments: the AI has profoundly changed how the human thinks (for similar results see Krügel et al 2023, Guo et al 2024; Kosmyna et al 2025, Vicente et al 2025).

[8] Their "hostile scaffolding" is about external elements whereas colonization is specifically about scaffolds that become constitutively integrated.

Second, we see **exogenous** directional **governance** in which the system's outputs are systematically shaped by optimization criteria set by external actors or infrastructures rather than by the agent's own reflective standpoint. This leads to a (third) normative issue as well: **misalignment**. One striking element is that AI possesses its own *directional properties*. Its outputs reflect training data distributions, optimization objectives, alignment procedures, and corporate design choices. That means that AI systems are far from value-neutral actors and introduce systematic biases that are extremely pervasive and not human aligned (Anwar et al 2024). Also, the design logic across the AI systems now functioning as System 0 is dominated by one objective: engagement (*The Economist*, 2026). What engagement means in practice is comfort, predictability, and the removal of friction. Riva et al. (2025) have a name for the trouble this creates: the comfort-growth paradox. Comfort-driven design is very good at what it is designed to do, which is keep users engaged; it is correspondingly bad at preserving the dissonance that learning and creativity require, and that Vygotsky (1978) long ago placed at the center of cognitive development. System 0, on its current trajectory, is thus an infrastructure oriented against growth. It delivers what the user already finds relevant, reinforcing existing cognitive patterns instead of pushing against them. None of this announces itself. It operates through what Pedwell (2024) calls algorithmic intuition aka patterns that feel as though they emerge from the user rather than being handed to her, and that narrow her cognitive range even as they appear to widen it. Misaligned, but experienced as native.

Fourth, **opacity**: there is a lot of work on the opacity of AI (Fisher et al 2015; Krügel et al 2023; Göndöcs et al 2025). System 0's optimization criteria are proprietary and legally shielded. Many AI models are also black boxes, whose computations and internal workings are difficult to penetrate.

Colonization denotes the installation of foreign governing structures within a territory such that the territory's own practices come to serve external interests (possibly while the inhabitants experience these processes as their own). Parallely, Cognitive Colonization is the incorporation of externally designed optimization objectives into the architecture of the self, such that the agent's pre-reflective processing has been shaped by interests that are not the agent's own.

This differs from epistemic deference (where the agent more or less knowingly relies on an external authority) and from manipulation (where the agent is nudged from outside). In Cognitive Colonization, the foreign objectives operate from *within* the cognitive system, below the threshold of reflective awareness, and are experienced as native. The structure is what makes it distinctively threatening: there is no external authority to refuse, no manipulation to detect, because the mechanism has become part of how the agent thinks.

This is fairly new in the history of cognitive extension. Notebooks, calculators, and maps are passive: they do not push back, do not have training objectives, and do not adapt their outputs based on engagement metrics (with caveats, see Liao and Huebner 2021). Language and culture have agendas (ideological, normative), but these are in principle inspectable rather than seamlessly integrated into the agent's pre-reflective processing. The novelty of the present condition is thus not extension alone, but opaque extension under exogenous directional governance. AI-as-System-0 combines the constitutive intimacy of a cognitive extension with the directional properties of an autonomous system whose biases and goals are structurally invisible to the user (Floridi, 2025). The human agent's deliberation and conditions have been co-constituted by a process with its own directional push, and the line between “our judgment” and “the AI’s output” gradually dissolves.

| Condition of Colonization | Type | Core claim | Contrast with offloading | Empirical / diagnostic marker |
|---|---|---|---|---|
| ***DESCRIPTIVE CONDITIONS — what kind of architectural situation the agent is in*** | | | | |
| **Exogenous directional governance / authorship** | Descriptive | The integrated layer's biases are set by criteria originating outside the agent rather than by the agent's own reflective standpoint. | Pure offloading outsources computation but mostly leaves direction to the agent (a calculator computes what the agent asks it to). Colonized S0 supplies both task completion and goals. | Systematic biases reflecting training data and corporate design (Anwar et al. 2024); engagement-maximization as stated business objective; proprietary optimization criteria. |
| **Downstream incorporation** | Descriptive | The resulting judgments, preferences, and habits are taken up by the agent *as her own* and persist beyond the moment of AI interaction aka they become part of how she reasons and perceives rather than remaining as acknowledged external inputs. | Offloading leaves the agent's unassisted judgment unchanged. Colonization reshapes the architecture so the directional bias carries over into contexts where the AI is absent. | Persistence after removal (Vicente & Matute 2023; (Salvi at al 2025). Bias amplification exceeding human-to-human baselines (Glickman & Sharot 2025). Vocabulary/creative-output homogenization (Hohenstein; Doshi & Hauser). |
| **Opacity** | Descriptive | The directional shaping and goals are not accessible to the agent at the point of use aka they cannot be identified as a distinct influence open to evaluation. The agent experiences the outputs, not the optimization criteria that produced them. | Offloaded tools are transparent about what they are (the calculator is a calculator). Colonized S0 operates invisibly and are inspectable by user or designer. | Users unaware of influence magnitude (Glickman & Sharot); breakdown-triggered metacognitive surprise when AI is disrupted (Bowker & Star inversion); inability to articulate what the system is optimizing for (Floridi 2025). |

| NORMATIVE CONDITION — why this architectural situation is a problem | | | | |
|---|---|---|---|---|
| **Misalignment with reflective endorsement** | Normative | The layer's outputs would fail the agent's reflective endorsement ( Moran 2001; Frankfurt 1971). | | Large literature on gap between AI-shaped salience/desirability patterns and reflectively avowed aims |

TABLE 3 – Description of System 0 as Cognitive Colonization

### *5.1 Beyond the Extended Mind*

We should address directly the possible theoretical tension our position creates within the extended mind framework (Clark & Chalmers, 1998). The classic extended mind thesis mostly assumed, implicitly, that cognitive extensions serve the agent's cognitive purposes. Otto's notebook records what Otto wants to remember, and the system functions as a whole in the service of Otto's goals. This is an architectural and functional claim and does not by itself guarantee that all constitutively integrated components will be normatively aligned with the agent's self-governing standpoint. In this way. System 0 does not contradict the extended mind thesis but it reveals a possibility that earlier formulations, developed in relation to passive artifacts, did not sufficiently explore the possibility of a normative (not functional) misalignment.[9] In particular, colonization occurs when this integration is normatively misaligned while functionally aligned. This is one of the framework's diagnostic contributions in so far as it makes visible a form of cognitive extension that prior philosophy of mind, developed in a world of passive cognitive tools, could not anticipate.

[9] Theorists have not ignored the ethical implications of extending the mind. See for instance Smart et al. 2017 and Clowes et al 2024.

We should equally address a second theoretical tension: if System 0 is constitutively part of the agent's own cognitive system, the worry is that the System 0 account appears to generate a paradox structurally similar to the one it identifies in Shaw and Nave (2026): if AI is part of us, how can it *threaten* us from within without that paradox undermining the claim that it is part of us?

A component threatens the agent when it systematically produces outputs that would fail reflective endorsement by that same agent (Moran, 2001; Frankfurt, 1971). In a sense, System 1 poses this threat routinely: implicit biases operate below reflective awareness and often fail the endorsement test (Kahneman, 2011; Stanovich, 2011). We do not conclude that System 1 is external to the agent, nor that being influenced by one's own System 1 constitutes "colonization". We say, rather, that the agent's own architecture contains components whose outputs can work against the agent's reflective interests, and we treat this as a structural feature of agency requiring internal monitoring and correction.

However, S1 is no System 0 either. The distinction that matters here is between what is *native* to a cognitive architecture situated in the brain, and what has been *integrated* into it. System 1 is native, in the sense that it grew with the agent, out of biological endowment, embodied history, and accumulated socialization. That means that its directional properties, whatever their flaws, are the agent's own in the sense that they belong to her body and complete identity. Not so with System 0, which joins an architecture that is already there, located in a particular type of implementation. This is what we mean by native here. Let's be clear: psychological Integration can be deep — deep enough to satisfy every condition the extended mind thesis asks of a cognitive extension — without ever making the integrated component *native*. And this is exactly where colonization lives. An integrated-but-not-native component can carry goals and norms the agent never authored, and can keep carrying them after integration is complete. That spatial difference is what the term colonization is meant to name: not an alien force pressing from outside, and not a native tendency the agent has to live with, rather something in between. A child who grows up with AI from the earliest years does integrate it deeply yet the integration is not native. Nativeness here is a claim about the substrate. A naturalized citizen is fully a citizen in every operative sense. Integration is complete even though the fact of naturalization does not disappear. Nothing about this diminishes the reality of their citizenship, it merely marks that the route in was different, and that different route leaves structural traces even after the status is settled. System 1, whatever its content, is realized in the biological brain and native as such. System 0 remains a *mind* extension, not a *brain* extension.

## 6. Conclusion

The three frameworks examined here (Tri-System Theory, Thinkframes, and System 0) are differently positioned with respect to three fundamental analytical dimensions: the locus of AI's operation (within or around the individual mind); the nature of AI's integration with the human cognitive system (constitutive versus merely causal); and the scope of explanatory reach (episodic individual events versus population-level dynamics versus both simultaneously). System 3 operates at the level of the individual agent engaged in a discrete reasoning episode with its central contribution being the experimental identification of cognitive surrender. Because System 3 presupposes a consultation event, we believe it struggles to accommodate AI's most pervasive contemporary mode: the preprocessing of informational environments that occurs before any consultation is initiated.

Thinkframes occupy the opposite end of the analytical spectrum. The framework's capacity to explain macro-level phenomena (e.g., convergence of interpretive frames across millions of users, the homogenization of collective attention) is genuine yet, though, by positioning AI as an ecological condition that individuals are embedded in, rather than as a process that individuals think with, Thinkframes does not capture the active, architecturally embedded role of AI (e.g., the continuously personalized preprocessing).

System 0 occupies the position alternative views have a hard time explaining: its architecture is simultaneously individual and collective, constitutive and ecological, episodic and continuous. At the individual level, System 0's constitutive integration account predicts the phenomena System 3 documents (high adoption rates, reduced critical evaluation). At the collective level, System 0's analysis of shared optimization logic (engagement maximization, comfort, frictionlessness instantiated across millions of individually calibrated interactions) predicts the phenomena Thinkframes documents (epistemic homogenization, convergent interpretive frames) as the population-level aggregate of individually instantiated processes. System 0 thus stands in a specific theoretical relation to both alternative views: it identifies the generative mechanism from which their downstream patterns emerge, and explains both individual and collective manifestations as products of the same underlying architecture.

Importantly, this paper's most original contribution - the concept of Cognitive Colonization - talks about: a constitutive integration, the installation of foreign optimization objectives within the architecture of cognition itself, below the threshold at which those objectives could be detected and evaluated. The three frameworks thus describe three distinct relations between the human agent and AI: System 3 describes an agent who defers to something outside herself; Thinkframes describes an agent who is shaped by something around her; System 0 describes an agent who has been, at least partially, reconstituted from within.

This tripartite distinction has direct implications for how we assess both the stakes of AI's cognitive impact and the interventions available to us. If the primary concern is cognitive surrender, the remedy is arguably epistemic vigilance: users need better critical thinking skills, stronger habits of independent verification, and more calibrated trust. If the primary concern is ecological homogenization, the remedy is environmental redesign: diversify the information ecosystem, regulate algorithmic curation, and build more pluralistic epistemic infrastructures. Both interventions remain valuable and necessary. But if the primary concern is Cognitive Colonization then neither vigilance nor environmental reform is sufficient on its own. The problem is, increasingly, constitutive of how the agent thinks. The goal is thus transformed, by the System 0 framework, from a technical safety challenge into a prerequisite for the integrity of human cognitive agency itself.